\documentclass[lettersize,journal]{IEEEtran}
\usepackage{amsmath,amsfonts}
\usepackage{algorithm}
\usepackage{array}
\usepackage[caption=false,font=normalsize,labelfont=sf,textfont=sf]{subfig}
\usepackage{textcomp}
\usepackage{stfloats}
\usepackage{url}
\usepackage{verbatim}
\usepackage{graphicx}
\usepackage{cite}
\newcommand{\eg}{\textit{e}.\textit{g}.}

\newcommand{\etal}{\textit{et~al.}}

\usepackage{algpseudocode}
\usepackage{amssymb}
\usepackage{float}
\usepackage{booktabs}
\usepackage{wrapfig}
\usepackage{microtype}
\usepackage{hyperref}
\usepackage{multirow}
\usepackage{makecell}
\usepackage{fontawesome5}

\usepackage[hypcap=false]{caption}

\usepackage[percent]{overpic}
\usepackage{xcolor}
\usepackage{tikz}

\begin{document}

\title{Can Video Diffusion Models Predict Past Frames? \\Bidirectional Cycle Consistency for Reversible Interpolation}

\author{~Lingyu~Liu,~Yaxiong Wang$^\dag$,~Li Zhu,~Zhedong~Zheng$^\dag$,~\IEEEmembership{Senior~Member,~IEEE}
\thanks{Lingyu Liu and Li Zhu are with the School of Software, Xi'an Jiaotong University, Xi'an, 710049, China. (e-mail: liulingyu@stu.xjtu.edu.cn; zhuli@xjtu.edu.cn).}
\thanks{Yaxiong Wang is with the School of Computer and Information Science, Hefei University of Technology, Jianghuai Advance Technology Center, Anhui Provincial Key Laboratory of Humanoid Robots. Anhui Provincial Industry Innovation Center of Humanoid Robots, Hefei, 230000, China. (e-mail: wangyx15@stu.xjtu.edu.cn).}
\thanks{Zhedong Zheng is with Faculty of Science and Technology, and Institute of Collaborative Innovation, University of Macau, Macau, 999078, China.   (e-mail: zhedongzheng@um.edu.mo).}
\thanks{$^\dag$ Corresponding author.}	}

\IEEEaftertitletext{
\vspace{-25pt}
\begin{center}
\textcolor{magenta}{
Project page: \href{https://lingyuliu.github.io/CVFI/}{https://lingyuliu.github.io/CVFI/}
}
\end{center}
}

\markboth{Journal of \LaTeX\ Class Files,~Vol.~14, No.~8, August~2021}%
{Shell \MakeLowercase{\textit{et al.}}: A Sample Article Using IEEEtran.cls for IEEE Journals}


\maketitle

\begin{abstract}
Video frame interpolation aims to synthesize realistic intermediate frames between given endpoints while adhering to specific motion semantics. While recent generative models have improved visual fidelity, they predominantly operate in a unidirectional manner, lacking mechanisms to self-verify temporal consistency. This often leads to motion drift, directional ambiguity, and boundary misalignment, especially in long-range sequences. Inspired by the principle of temporal cycle-consistency in self-supervised learning, we propose a novel bidirectional framework that enforces symmetry between forward and backward generation trajectories. Our approach introduces learnable directional tokens to explicitly condition a shared backbone on temporal orientation, enabling the model to jointly optimize forward synthesis and backward reconstruction within a single unified architecture. This cycle-consistent supervision acts as a powerful regularizer, ensuring that generated motion paths are logically reversible. Furthermore, we employ a curriculum learning strategy that progressively trains the model from short to long sequences, stabilizing dynamics across varying durations. Crucially, our cyclic constraints are applied only during training; inference requires a single forward pass, maintaining the high efficiency of the base model. Extensive experiments show that our method achieves state-of-the-art performance in imaging quality, motion smoothness, and dynamic control on both 37-frame and 73-frame tasks, outperforming strong baselines while incurring no additional computational overhead. The videos are available on our project website.
\end{abstract}


\begin{figure*}[t]
 \centering
  \vspace{-.1in}
 \includegraphics[width=0.90\linewidth]{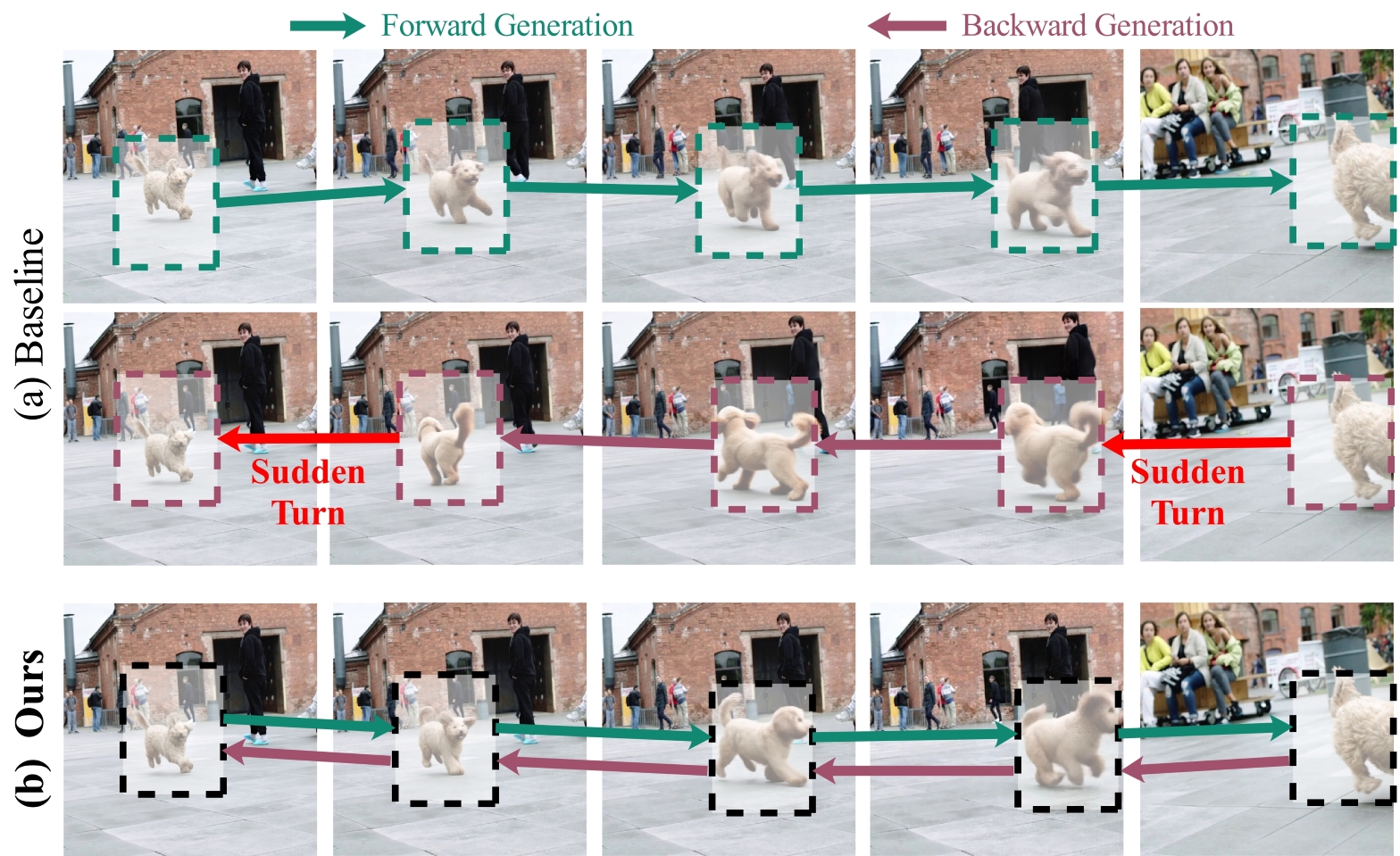}
 \caption{Cycle-consistency of Time. Given identical start/end frames, we test temporal symmetry by generating (top) a forward sequence and (bottom) its time-reversed counterpart via swapped endpoints. The baseline fails to synthesize true backward motion and instead resolves the constraint via a directional flip, where the dog re-orients to walk forward. In contrast, our model achieves robust cycle-consistency. It captures authentic reverse dynamics, producing a coherent backward-walking video that preserves the subject's orientation, thereby closing the temporal loop faithfully.}
 \label{fig:cct}
 \vspace{-.1in}
\end{figure*}

\section{Introduction}
\label{sec:intro}
\IEEEPARstart{V}{ideo} Frame Interpolation (VFI) aims to synthesize temporally coherent and visually plausible intermediate frames between a given start and end image. This task is fundamental to many applications, including slow-motion video generation~\cite{zhang2025eden}, frame-rate conversion~\cite{framer,zhang2026arbitrary,lu2025self}, cinematic post-production~\cite{pardo2025matchdiffusion}, and interactive content creation~\cite{sfi}. Recently, the integration of textual guidance has further enabled users to specify desired motion semantics, giving rise to text-guided video interpolation. Despite significant advances in generative modeling, producing high-fidelity interpolations that maintain long-range temporal consistency, respect boundary conditions, and align with complex motion descriptions remains an open challenge.

Existing approaches generally fall into two categories. Traditional optical flow-based approaches~\cite{ding2024video,prasanna2024video} rely on explicit motion estimation to warp and blend frames. While effective for small displacements, they often fail under large motions, occlusions, or non-rigid deformations, resulting in artifacts such as blurring, tearing, or ghosting. On the generative side, while diffusion-based models leverage strong priors for high-fidelity synthesis, they predominantly operate in a unidirectional manner.
Some recent works~\cite{framer,generativeInbetweening,yangvibidsampler,li2021deep} attempt to incorporate backward generation but typically train separate models or adapters for reversed directions without enforcing joint consistency constraints. Consequently, these models lack a mechanism to self-verify whether the generated forward trajectory is logically reversible, leading to motion drift or semantic ambiguity in long sequences. Moreover, these techniques are almost exclusively built on Stable Diffusion backbones and constrained to fixed-length outputs, most commonly 25 frames, which limits their applicability to longer and more complex motions. In contrast, modern long-video generative models~\cite{wan,hunyuanvideo,framepacking} can produce videos of variable lengths, but they have not been specifically trained for video frame interpolation tasks. Although some training-free strategies can be employed to perform interpolation, these approaches may fail to perfectly align with the target end frame when handling complex scenarios. 

Inspired by the principle of temporal cycle-consistency in self-supervised representation learning~\cite{cycle-consistency}, we hypothesize that a robust video generative model should produce dynamics that are consistent under time reversal. Specifically, if a model can generate a plausible forward sequence from frame A to B, then applying the same model with reversed temporal context should yield a coherent backward sequence from B to A. As shown in Figure~\ref{fig:cct}, we provide the start and end frames of a dog moving forward and ask the model to generate a forward-walking video. When we swap the start and end frames and request the model to generate a backward-walking video, the untrained baseline model produces a video where the dog turns around and walks forward again, whereas our model successfully generates a video of the dog walking backward.
We propose that explicitly modeling this bidirectional loop within a single framework provides a powerful self-supervised signal to regularize motion learning.

To realize this, we introduce a cycle-consistent training strategy for text-guided video frame interpolation. The model is supervised to minimize reconstruction errors in both temporal directions simultaneously. 
We incorporate two learnable text tokens that act as explicit directional cues, allowing the model to condition generation on either forward or backward temporal orientation. 
To handle the challenges of long-range coherence, we employ a curriculum learning strategy: we first train on 37-frame sequences to stabilize short-term dynamics and then fine-tune on 73-frame sequences to capture extended motion patterns. Crucially, our cyclic supervision is applied only during training; inference requires only a single forward pass, ensuring no additional computational overhead.
Our key contributions are as follows:
\begin{itemize}
\item We propose a bidirectional cycle-consistent framework for video frame interpolation that enforces temporal symmetry via learnable directional tokens and joint forward-backward optimization in a unified architecture.  We also introduce a curriculum learning schedule that progressively extends training from short to long sequences, ensuring robust dynamics across varying durations.
\item Extensive experiments show that our method is complementary to existing approaches, yielding state-of-the-art results in visual quality, motion smoothness, and long-range temporal consistency (\eg, on 37- and 73-frame tasks), while outperforming prior art with no extra computational cost at inference.
\end{itemize}

\section{Related Work}
\noindent\textbf{Video Frame Interpolation.}
Traditional video frame interpolation generates intermediate frames via optical flow-based alignment~\cite{bao2019depth,niklaus2017video,hu2024iq,liu2024sparse,huang2022real,reda2022film,li2023amt,zhang2023extracting}, yet often suffers from artifacts under large motions or occlusions. Leveraging generative priors from diffusion models~\cite{ho2022video,yang2023diffusion,svd,cogvideox,blattmann2023align}, recent methods achieve superior inbetweening~\cite{chen2025sci,danier2024ldmvfi,shen2024dreammover,voleti2022mcvd,huang2024motion}. For instance, Li~\etal~\cite{li2021deep} use sketch-conditioned bidirectional generation for cartoon inbetweening, while Lu~\etal~\cite{lu2025self} employ event-guided self-supervision to correct rolling shutter distortions without ground truth.
Furthermore, several approaches enable full video interpolation from start and end frames~\cite{versa,generativeCondition,jain2024video,xing2024tooncrafter,xing2024dynamicrafter}. Feng~\etal~\cite{feng2024explorative} propose training-free bounded generation for controllable interpolation; Wang~\etal~\cite{generativeInbetweening} leverage backward-in-time denoising for coherent synthesis; and Jeon~\etal~\cite{jeon2026motion} align forward-backward motion priors during time-reversal sampling. However, these reverse-guided methods require repeated noising and sampling, significantly increasing inference time.

\noindent\textbf{Video Generation Models.}
Recent progress in video generation~\cite{lu2026anifeats,xu2026anchorcrafter,zhu2025zero,yang2026multipaint} has enabled diverse controllable synthesis tasks. Qu~\etal~\cite{qu2025controllable} achieve sketch-conditioned human video generation with temporal consistency; Kang~\etal~\cite{kang2026drivegen} enable multi-view autonomous driving video generation via shared geometry-temporal encoding; and Xing~\etal~\cite{xing2024make} integrate text, structure, and concept adapters for precise customization.
For long-form generation, models like Open-Sora~\cite{Open-Sora} and HunyuanVideo~\cite{hunyuanvideo} use efficient attention and frame packing but are designed for open-ended text-to-video synthesis, lacking explicit endpoint interpolation control and prone to temporal drift even with latent conditioning. SFI~\cite{sfi} adapts WAN~\cite{wan} with length-specific LoRA adapters, yet lacks bidirectional control. To our knowledge, our work is the first to adapt long video generators for bidirectional interpolation via directional prompting and symmetric training, bridging controllable VFI and scalable generative modeling.

\section{Methodology}
\begin{figure*}[!t]
 \centering
  \vspace{-.1in}
 \includegraphics[width=0.98\linewidth]{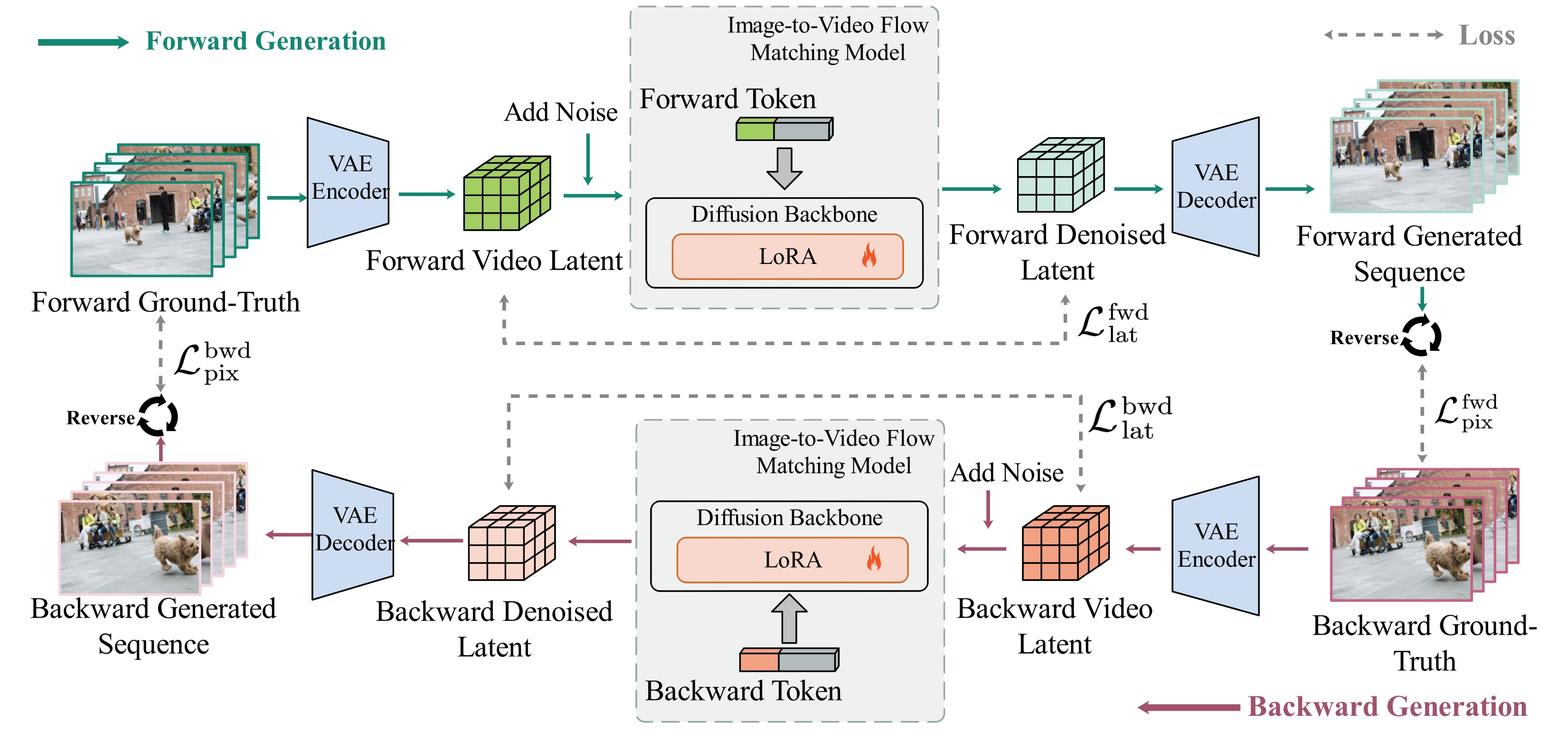}
  \vspace{-.1in}
 \caption{A brief overview of our framework. During training, each ground-truth video is used to construct two samples. The forward sample interpolates from the original start frame to the original end frame. The backward sample interpolates in the reverse temporal direction, starting from the original end frame and ending at the original start frame. These two directions are controlled by distinct learnable directional tokens. The model is supervised with reconstruction losses in both latent space and pixel space for both directions, which encourages consistent motion modeling under time reversal.}
 \label{fig1}
  \vspace{-.1in}
\end{figure*}

We present a bidirectional framework for text-guided video frame interpolation that leverages \textit{temporal cycle-consistency} as a structural prior to enhance motion coherence. Built upon a pre-trained long-video generative model based on Rectified Flow, our approach enforces symmetry between forward and backward generation. As illustrated in Figure~\ref{fig1}, the core idea is that a robust interpolation model should not only generate a plausible sequence from a start frame $I_1$ to an end frame $I_L$, but also be capable of reversing this process to reconstruct $I_1$ from $I_L$ using the same parameters. This cyclic constraint acts as a powerful self-supervised signal to regularize motion learning, particularly over extended durations. The overall pipeline consists of a shared flow-based backbone and a lightweight directional conditioning mechanism, ensuring high fidelity without compromising inference efficiency.

\subsection{Preliminary: Image-to-Video Flow Matching Model}
Our method is scalable to various video backbones. To simplify the illustration, we take  FramePack~\cite{framepacking}, a long-video generative model derived from HunyuanVideo~\cite{hunyuanvideo} that employs Rectified Flow instead of traditional diffusion, as an example. Rectified Flow learns a deterministic ordinary differential equation (ODE) that transports the data distribution to a Gaussian noise distribution along straight paths, enabling faster convergence and more stable sampling.

Given a video sequence $\mathcal{V} = \{I_1, I_2, \dots, I_L\}$ and a text prompt, FramePack operates in the VAE latent space. The video is encoded into a latent tensor $x \in \mathbb{R}^{b \times c \times l \times h \times w}$, where $b$ is the batch size, $c$ is the latent channel dimension, $l$ denotes the temporal length, and $(h, w)$ are the spatial dimensions after downsampling.
The model is trained to predict the velocity field $v(x_t, t)$ at each time step $t \in [0, 1]$, where $x_t$ is the interpolated latent state along the path from clean data $x_0$ to noise $\epsilon$. The training target is defined as $v_{\text{target}} = \epsilon - x_0$, where $\epsilon \sim \mathcal{N}(0, I)$ is a random Gaussian vector. To support arbitrary-length generation, FramePack employs a multi-stage frame packing mechanism that compresses historical frames into compact visual tokens, keeping the transformer context length bounded. During inference, frames are generated autoregressively by integrating the learned ODE forward in time. FramePack can be adapted for video frame interpolation in a training-free manner by injecting the latent representation of the target end frame into its frame packing mechanism. However, the model remains inherently unidirectional, as it is trained exclusively for forward generation and thus lacks mechanisms to verify temporal consistency, which often leads to motion drift over long durations.

\subsection{Directional Conditioning Mechanism}
To mitigate the unidirectional bias inherent in long-range interpolation, we propose a bidirectional training framework that leverages time-reversal symmetry as a structural regularizer. 
Let $p$ denote the input text prompt and $E_{\text{text}}(\cdot)$ be the pre-trained text encoder. The original semantic condition is obtained as $f_{\text{text}} = E_{\text{text}}(p) \in \mathbb{R}^{L_p \times d}$, where $L_p$ is the sequence length of the text tokens and $d$ is the hidden dimension.
To explicitly control the temporal orientation without modifying the backbone weights, we introduce two learnable directional tokens, $\tau_{\text{fwd}}, \tau_{\text{bwd}} \in \mathbb{R}^{d}$. We construct the final conditioning sequence $f_{\text{cond}}$ by concatenating the specific directional token with the semantic embedding:
\begin{equation}
    f_{\text{cond}}^{(d)} = \tau_d \oplus f_{\text{text}}, \quad \text{where } d \in \{\text{fwd}, \text{bwd}\},
\end{equation}
where $\oplus$ denotes the concatenation operation along the sequence dimension. 
This augmented condition $f_{\text{cond}}^{(d)}$ is then fed into the Diffusion Transformer (DiT) backbone. Consequently, the velocity field prediction $v_\theta$ becomes conditional on the direction:
\begin{equation}
    v_\theta(x_t, t, f_{\text{cond}}^{(d)}) = 
    \begin{cases} 
      v_\theta(x_t, t, \tau_{\text{fwd}} \oplus f_{\text{text}}) & \text{(fwd: } I_1 \to I_L), \\
      v_\theta(x_t, t, \tau_{\text{bwd}} \oplus f_{\text{text}}) & \text{(bwd: } I_L \to I_1).
    \end{cases}
\end{equation}
By sharing the backbone parameters $\theta$ while varying only the prefix token $\tau_d$, the model learns to align its motion manifold with the specified temporal flow, effectively acting as a switch between forward and reverse dynamics.

\subsection{Cycle-Consistent Bidirectional Training}
We supervise both directions by minimizing the reconstruction error across both latent and pixel spaces. Let $\hat{x}_{0,d}$ denote the predicted clean latent for direction $d \in \{\text{fwd}, \text{bwd}\}$, and let $x_d^*$ and $\mathcal{V}_d^*$ be the corresponding ground-truth latent and pixel sequences. The total objective is formulated as the sum of four distinct consistency terms:
\begin{equation}
\begin{aligned}
\mathcal{L} _{\mathrm{total}}=\;\underbrace{\parallel \hat{x}_{0,\mathrm{fwd}}-x_{\mathrm{fwd}}^{*}\parallel _{2}^{2}}_{\mathcal{L} _{\mathrm{lat}}^{\mathrm{fwd}}}+\underbrace{\parallel \mathcal{D} (\hat{x}_{0,\mathrm{fwd}})-\mathcal{V} _{\mathrm{fwd}}^{*}\parallel _{2}^{2}}_{\mathcal{L} _{\mathrm{pix}}^{\mathrm{fwd}}}
\\
+\lambda _{rev}\left( \underbrace{\parallel \hat{x}_{0,\mathrm{bwd}}-x_{\mathrm{bwd}}^{*}\parallel _{2}^{2}}_{\mathcal{L} _{\mathrm{lat}}^{\mathrm{bwd}}}+\underbrace{\parallel \mathcal{D} (\hat{x}_{0,\mathrm{bwd}})-\mathcal{V} _{\mathrm{bwd}}^{*}\parallel _{2}^{2}}_{\mathcal{L} _{\mathrm{pix}}^{\mathrm{bwd}}} \right) ,
\end{aligned} \label{eq:main_loss}
\end{equation}
where $\mathcal{D}(\cdot)$ denotes the VAE decoder and $\mathcal{V} _{\mathrm{bwd}}^{*}=\mathrm{Reverse(}\mathcal{V} _{\mathrm{fwd}}^{*})$. The hyperparameter $\lambda _{rev}$ balances the contribution of the backward reconstruction losses relative to the forward ones. In all our experiments, we set $\lambda _{rev} = 1$ by default, treating forward and backward directions with equal importance to enforce symmetric cycle-consistency. By optimizing this quadruple objective, the model learns a velocity field $v_\theta(x_t, t, f_{\text{cond}}^{(d)})$ that remains consistent under time reversal, effectively regularizing the motion manifold to prevent drift and ensure strict boundary alignment.

\noindent\textbf{Relation to cycle consistency.}
In frame interpolation, both endpoints $(I_1, I_L)$ are given as boundary conditions, so cycle consistency does not concern returning to the starting frame, which is satisfied by construction, but rather the agreement between the two generated \emph{trajectories}: a temporally symmetric model should produce a forward sequence that coincides with the time-reversal of its backward sequence. Let $\hat{\mathcal{V}}_{\mathrm{fwd}} = \mathcal{D}(\hat{x}_{0,\mathrm{fwd}})$ and $\hat{\mathcal{V}}_{\mathrm{bwd}} = \mathcal{D}(\hat{x}_{0,\mathrm{bwd}})$ denote the decoded forward and backward sequences. We define the cycle discrepancy as:
\begin{equation}
\mathcal{C} \;=\; \big\| \hat{\mathcal{V}}_{\mathrm{fwd}}
- \mathrm{Reverse}(\hat{\mathcal{V}}_{\mathrm{bwd}}) \big\|_2 .
\label{eq:cycle}
\end{equation}
Although $\mathcal{C}$ is not explicitly optimized, it is
upper-bounded by our training objective. Since
$\mathrm{Reverse}(\cdot)$ is a permutation of temporal indices and
hence norm-preserving, and $\mathcal{V}^{*}_{\mathrm{bwd}} =
\mathrm{Reverse}(\mathcal{V}^{*}_{\mathrm{fwd}})$ by construction,
the triangle inequality gives
\begin{equation}
\begin{aligned}
\mathcal{C}
&\le \big\| \hat{\mathcal{V}}_{\mathrm{fwd}}
 - \mathcal{V}^{*}_{\mathrm{fwd}} \big\|_2
 + \big\| \mathrm{Reverse}(\hat{\mathcal{V}}_{\mathrm{bwd}})
 - \mathcal{V}^{*}_{\mathrm{fwd}} \big\|_2 \\
&= \big\| \hat{\mathcal{V}}_{\mathrm{fwd}}
 - \mathcal{V}^{*}_{\mathrm{fwd}} \big\|_2
 + \big\| \hat{\mathcal{V}}_{\mathrm{bwd}}
 - \mathcal{V}^{*}_{\mathrm{bwd}} \big\|_2 \\
&= \sqrt{\mathcal{L}^{\mathrm{fwd}}_{\mathrm{pix}}}
 + \sqrt{\mathcal{L}^{\mathrm{bwd}}_{\mathrm{pix}}},
\end{aligned}
\label{eq:bound}
\end{equation}
or equivalently $\mathcal{C}^2 \le
2\,(\mathcal{L}^{\mathrm{fwd}}_{\mathrm{pix}}
+ \mathcal{L}^{\mathrm{bwd}}_{\mathrm{pix}})$. An identical argument
holds in the latent space for $\hat{x}_{0,\mathrm{fwd}}$ and
$\hat{x}_{0,\mathrm{bwd}}$ with the latent reconstruction terms
$\mathcal{L}^{\mathrm{fwd}}_{\mathrm{lat}}$ and
$\mathcal{L}^{\mathrm{bwd}}_{\mathrm{lat}}$, using
$x^{*}_{\mathrm{bwd}} = \mathrm{Reverse}(x^{*}_{\mathrm{fwd}})$.
Minimizing Eq.~\eqref{eq:main_loss} thus implicitly drives the cycle
discrepancy toward zero: the bidirectional reconstruction losses
jointly constrain the forward and time-reversed backward trajectories
to agree, without requiring an expensive two-stage
generate-then-reverse procedure during training. In practice, we
report the per-pixel mean absolute error between
$\hat{\mathcal{V}}_{\mathrm{fwd}}$ and
$\mathrm{Reverse}(\hat{\mathcal{V}}_{\mathrm{bwd}})$, denoted
$\mathcal{C}_{\ell_1}$, as an empirical measure of
Eq.~\eqref{eq:cycle}; by the equivalence of finite-dimensional norms,
controlling the $\ell_2$ bound in Eq.~\eqref{eq:bound} likewise
controls $\mathcal{C}_{\ell_1}$. In
Section~\ref{sec:exp_cycle}, we verify that this implicit supervision
substantially reduces the inference-time cycle discrepancy relative
to the unidirectional baseline.

\noindent\textbf{Discussion. Why bidirectional training?} The intuition behind bidirectional training is to enforce temporal consistency and reversibility as an inductive bias for learning physically plausible dynamics. By jointly minimizing reconstruction errors in both forward and backward directions across latent and pixel spaces, we constrain the velocity field to lie on a coherent motion manifold where trajectories are invertible and drift-resistant. 
The backward pass serves as a strong regularizer by enforcing trajectory-level cycle consistency: the forward sequence and the time-reversal of the backward sequence, generated from the same endpoints with shared parameters, are jointly driven toward the same ground-truth trajectory and are therefore constrained to agree with each other (Eq.~\eqref{eq:bound}). A model that fits the forward direction with temporally incoherent shortcuts cannot simultaneously fit its time-reversed counterpart, which anchors the dynamics.
This symmetric objective, combined with multi-scale supervision at both latent and pixel levels, encourages the model to capture underlying physical dynamics rather than directional artifacts. It further mitigates error accumulation in long-term prediction and stabilizes optimization through complementary gradients. Ultimately, bidirectional training transforms the learning objective from simple forward prediction to learning a reversible mapping that respects temporal symmetry, leading to more robust generalization and physically grounded representations.

\subsection{Curriculum Learning Strategy}
\noindent\textbf{Multi-Rate Temporal Sampling.}
To support interpolation of arbitrary duration, our training data includes multiple temporal resamplings of the same video content at different playback rates. Specifically, for each source clip, we generate two variants: a short video of 37 frames and a long video of 73 frames. Both share identical start and end frames but exhibit different motion velocities across the sequence. This design encourages the model to generate videos at different frame rates, better aligning with the requirements of video frame interpolation tasks.

\noindent\textbf{Progressive Curriculum Learning.}
Furthermore, we adopt a curriculum learning strategy in which training begins on short videos, which provide a temporally compressed view of motion and allow the model to rapidly learn stable short-range transitions under bidirectional supervision. After convergence, we switch to the long videos, which require stronger long-range coherence. This progressive exposure to varying motion rates enables the model to generalize across interpolation lengths, producing temporally plausible videos that naturally accelerate or decelerate according to the specified number of output frames.

\noindent\textbf{Inference Efficiency.}
A key advantage of our framework is its computational efficiency. While the cycle-consistent supervision involves a dual-path structure during training, the inference phase requires only a single forward pass. Given a user-specified direction (typically forward), the model utilizes the corresponding token and generates the sequence autoregressively using the learned ODE solver. No backward generation or iterative refinement is needed at test time. Consequently, our method achieves state-of-the-art interpolation quality with latency identical to the base model, making it highly practical for real-world applications.

\noindent\textbf{Discussion. Why not train directly on long videos?} A naive end-to-end training on target interpolation lengths (\eg, 73 frames) presents two fundamental challenges. First, the model tends to learn a trivial shortcut: copying either the first or last frame throughout the entire sequence. This occurs because generating 73 coherent frames from scratch is an ill-posed problem in the early training stage, the temporal gradients are too sparse, and the long-range dependencies are too complex for the model to capture meaningful motion patterns. Consequently, it falls into a locally optimal but semantically empty solution: a still video that perfectly matches the start and end frames but lacks any motion. Second, the convergence is inherently unstable. Without a proper motion prior, the bidirectional supervision signals from distant frames often conflict or cancel out, leading to oscillating losses and slow convergence. The model struggles to disentangle whether a prediction error stems from local motion inaccuracies or global temporal inconsistency. Our progressive curriculum learning addresses both issues. By starting with short videos (37 frames), we provide dense temporal gradients that allow the model to first establish basic bidirectional motion dynamics. This phase acts as a warm-up, teaching the model that frames should change smoothly over time rather than remain static. Once this motion prior is internalized, we gradually increase the temporal horizon to 73 frames, enabling the model to build long-range coherence upon a solid foundation, without falling into the still-video trap.

\begin{figure*}[p]
 \centering
 \includegraphics[width=1\linewidth]{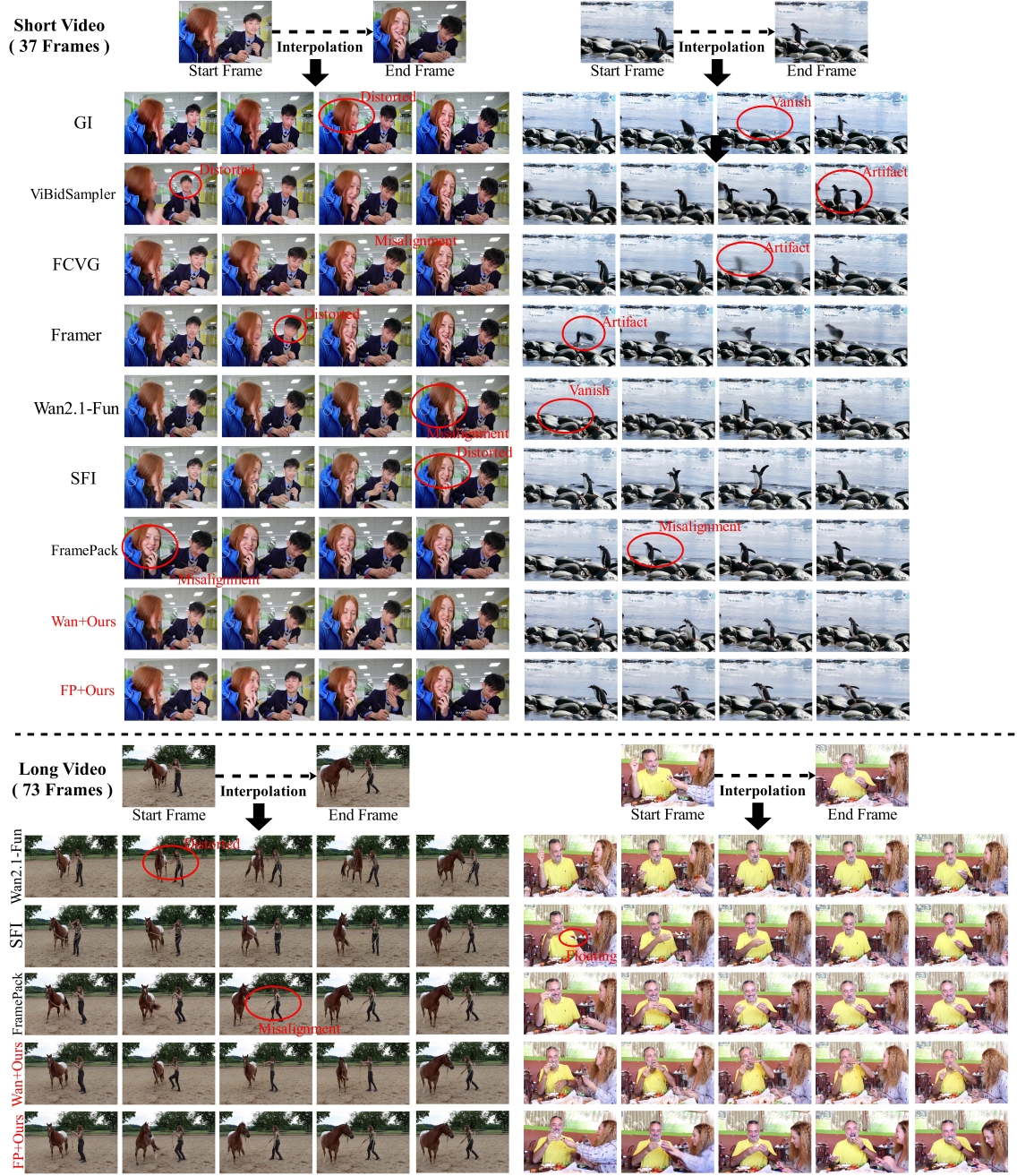}
 \caption{Qualitative Comparisons with Baselines. Our methods (\textit{Wan+Ours} and \textit{FP+Ours}) achieve significantly smoother trajectories and coherent temporal dynamics on both short videos (37 frames) and long videos (73 frames). Videos can be viewed in our project website.}
 \label{fig:baseline}
\end{figure*}

\section{Experiment}
\subsection{Implementation Details}
\noindent\textbf{Datasets.} We sample 5,000 videos from VidGen-1M~\cite{Vidgen-1m} as our training set based on three VBench metrics (subject consistency, motion smoothness, and dynamic degree) to ensure stable dynamics and persistent identities. The clips cover diverse motions and are preprocessed into 37-frame and 73-frame sequences with aligned endpoints. For evaluation, we construct two complementary test beds: (1) 100 non-overlapping videos from VidGen-1M for in-domain assessment, and (2) zero-shot evaluation on UltraVideo~\cite{ultravideo}, which remains unseen during training and features substantially different scenes and dynamics, providing a rigorous test of generalization.

\noindent\textbf{Settings.} 
We freeze all parameters of the original backbone model and train only two sets of newly introduced components: LoRA modules with rank 64 applied to the transformer layers, and two learnable directional tokens ($\tau_{\text{fwd}}$ and $\tau_{\text{bwd}}$). We use separate learning rates for these components: $2 \times 10^{-4}$ for the LoRA parameters and $2 \times 10^{-3}$ for the directional tokens. Training is performed for 4 epochs on the training set using the AdamW optimizer with a global batch size of 6, distributed across NVIDIA RTX PRO 6000 GPUs. The entire training process takes approximately three days.

\noindent\textbf{Evaluation Metrics.} Following most related works~\cite{sfi,chen2025sci}, we utilize FVD~\cite{fvd} and VBench~\cite{vbench} to measure the overall quality of the test videos. For a generated video with $N$ frames, we uniformly sample $N$ frames from the ground truth to compute FVD, ensuring matched frame counts across all methods. We select six relevant dimensions from VBench: Subject Consistency, Aesthetic Quality, Imaging Quality, Temporal Flickering, Motion Smoothness, and Dynamic Degree. 

\subsection{Comparison with Competitive Methods}
\noindent\textbf{Competitive Methods.} We conduct comparisons between our method and seven video frame interpolation methods: GI~\cite{generativeInbetweening},ViBidSampler~\cite{yangvibidsampler}, FCVG~\cite{generativeCondition}, Framer~\cite{framer}, Wan2.1-Fun~\cite{wan}, SFI~\cite{sfi}, and FramePack~\cite{framepacking}. GI, ViBidSampler, FCVG, and  Framer are generative models built upon Stable Video Diffusion~\cite{svd} with 1.5B parameters, limited to producing fixed-length sequences of 25 frames. Wan2.1-Fun is a lightweight variant of Wan2.1 with 1.3B parameters, specifically optimized for interpolation tasks. SFI utilizes the large-scale Wan2.1 backbone comprising 14B parameters and supports output lengths of 33 and 65 frames. FramePack leverages a 13B parameter HunyuanVideo~\cite{hunyuanvideo} backbone and produces videos of length $N \times 36 + 1$ through its autoregressive frame packing mechanism. Our primary implementation adopts the FramePack architecture. FP+finetuned denotes the FramePack backbone fine-tuned on the same data volume as our full model but using only forward videos.
By conditioning on both the start and end frames, this framework is adapted to perform interpolation at compatible lengths, specifically 37 and 73 frames, which serve as our main evaluation settings. 
To further validate the generalizability of our approach beyond the FramePack backbone, we also train an additional version of our method using the Wan2.1-Fun architecture.
For a fair comparison, all videos are generated at a resolution of $768 \times 512$ during evaluation.

\noindent\textbf{Quantitative Comparisons.}
Table~\ref{tab:base} presents the quantitative comparison across different sequence lengths on VidGen-1M, while Table~\ref{tab:base2} presents the corresponding numbers on UltraVideo.
To provide a fair and holistic assessment, we report FVD alongside six VBench perceptual metrics, with Ground‑Truth scores listed as the upper reference bound. We argue that superior performance should not be judged solely by extreme values on individual metrics, but by overall balance and proximity to the real video distribution.
Our two models, \textit{Wan+Ours} built upon Wan2.1-Fun and \textit{FP+Ours} built upon FramePack, achieve substantial improvements over their respective backbones on most metrics.

\begin{table*}[htbp]
  \centering
  \caption{Quantitative Comparisons with Baselines on VidGen-1M. Results are reported using FVD (lower is better) and six VBench metrics (higher is better). \textit{FP+finetuned} denotes the FramePack backbone fine-tuned on the same data volume as our full model but using only forward videos. Our full models achieve the highest average VBench scores across both 37-frame and 73-frame tasks, closely matching the Ground-Truth distribution and outperforming all baselines in imaging quality and motion smoothness.}
    \begin{tabular}{c|c|c|c|c|c|c|c|c|c|c}
    \toprule
    Method & \makecell{Backbone \\ Params} & Frames & FVD$\downarrow$   & \makecell{Subject \\ Consistency $\uparrow$} & \makecell{Aesthetic \\ Quality $\uparrow$} & \makecell{Imaging \\ Quality $\uparrow$} & \makecell{Temporal \\ Flickering $\uparrow$} & \makecell{Motion \\ Smoothness $\uparrow$} & \makecell{Dynamic \\ Degree $\uparrow$} & \makecell{VBench \\ Mean $\uparrow$} \\
    \midrule
    \textcolor[rgb]{ .502,  .502,  .502}{Ground-Truth} & \textcolor[rgb]{ .502,  .502,  .502}{-} & \textcolor[rgb]{ .502,  .502,  .502}{37} & \textcolor[rgb]{ .502,  .502,  .502}{-} & \textcolor[rgb]{ .502,  .502,  .502}{0.884 } & \textcolor[rgb]{ .502,  .502,  .502}{0.441} & \textcolor[rgb]{ .502,  .502,  .502}{0.636} & \textcolor[rgb]{ .502,  .502,  .502}{0.949} & \textcolor[rgb]{ .502,  .502,  .502}{0.987} & \textcolor[rgb]{ .502,  .502,  .502}{0.98} & \textcolor[rgb]{ .502,  .502,  .502}{0.81} \\
    GI    & 1.5B  & 25    & 698   & \textbf{0.921} & 0.424  & 0.593  & 0.966  & 0.985  & 0.63  & 0.75  \\
    ViBidSampler & 1.5B  & 25    & \textbf{693} & 0.863  & 0.419  & 0.540  & 0.944  & 0.962  & 0.96  & 0.78  \\
    FCVG  & 1.5B  & 25    & 992   & 0.906  & 0.442  & 0.554  & 0.967  & 0.989  & 0.63  & 0.75  \\
    Framer & 1.5B  & 25    & 806   & 0.882  & 0.389  & 0.525  & 0.961  & 0.980  & 0.85  & 0.76  \\
    Wan2.1-Fun (Wan) & 1.3B  & 37    & 984   & 0.876  & 0.451  & 0.584  & 0.948  & 0.965  & 0.96  & 0.80  \\
    \textbf{Wan+Ours} & 1.3B  & 37    & 781   & 0.891  & 0.450  & 0.597  & 0.957  & 0.977  & 0.96  & \textcolor[rgb]{ .753,  0,  0}{\textbf{0.81}} \\
    SFI   & 14B   & 33    & 910   & 0.879  & 0.454  & 0.561  & 0.950  & 0.974  & \textbf{1.00} & 0.80  \\
    FramePack (FP) & 13B   & 37    & 1049  & 0.897  & \textbf{0.479} & 0.585  & 0.981  & 0.992  & 0.83  & 0.79  \\
    FP+finetuned & 13B   & 37    & 932   & 0.880  & 0.461  & 0.553  & \textbf{0.988}  & 0.991  & 0.59  & 0.74  \\
    \textbf{FP+Ours} & 13B   & 37    & 885   & 0.885  & 0.465  & \textbf{0.602} & \textbf{0.988} & \textbf{0.992} & 0.90  & \textcolor[rgb]{ .753,  0,  0}{\textbf{0.81}} \\
    \midrule
    \textcolor[rgb]{ .502,  .502,  .502}{Ground-Truth} & \textcolor[rgb]{ .502,  .502,  .502}{-} & \textcolor[rgb]{ .502,  .502,  .502}{73} & \textcolor[rgb]{ .502,  .502,  .502}{-} & \textcolor[rgb]{ .502,  .502,  .502}{0.883} & \textcolor[rgb]{ .502,  .502,  .502}{0.454} & \textcolor[rgb]{ .502,  .502,  .502}{0.641} & \textcolor[rgb]{ .502,  .502,  .502}{0.960} & \textcolor[rgb]{ .502,  .502,  .502}{0.981} & \textcolor[rgb]{ .502,  .502,  .502}{0.97} & \textcolor[rgb]{ .502,  .502,  .502}{0.81} \\
    Wan2.1-Fun (Wan) & 1.3B  & 73    & 627   & 0.881  & 0.450  & 0.579  & 0.956  & 0.971  & \textbf{0.97} & 0.80  \\
    \textbf{Wan+Ours} & 1.3B  & 73    & \textbf{563} & \textbf{0.887} & 0.455  & 0.605  & 0.968  & 0.983  & 0.93  & 0.80  \\
    SFI   & 14B   & 65    & 622   & 0.877  & 0.457  & 0.562  & 0.977  & 0.980  & 0.93  & 0.80  \\
    FramePack (FP) & 13B   & 73    & 686   & 0.886  & \textbf{0.482} & 0.596  & 0.985  & 0.984  & 0.79  & 0.79  \\
    FP+finetuned & 13B   & 73   & 706   & 0.875  & 0.473  & 0.575  & \textbf{0.989}  & 0.986  & 0.36  & 0.71  \\
    \textbf{FP+Ours} & 13B   & 73    & 601   & 0.882  & 0.477  & \textbf{0.632} & 0.987 & \textbf{0.992} & 0.87  & \textcolor[rgb]{ .753,  0,  0}{\textbf{0.81}} \\
    \bottomrule
    \end{tabular}%
  \label{tab:base}%
\end{table*}%

\begin{table*}[htbp]
  \centering
  \caption{Zero-Shot quantitative evaluation on the unseen UltraVideo dataset. The results highlight the robust generalization of our method across different video distributions. Our methods achieve state-of-the-art scores in \textit{Motion Smoothness} and \textit{Dynamic Degree}, confirming their ability to generate coherent and dynamic videos even in zero-shot scenarios.}
    \begin{tabular}{c|c|c|c|c|c|c|c|c|c|c}
    \toprule
    Method & \makecell{Backbone \\ Params} & Frames & FVD$\downarrow$   & \makecell{Subject \\ Consistency $\uparrow$} & \makecell{Aesthetic \\ Quality $\uparrow$} & \makecell{Imaging \\ Quality $\uparrow$} & \makecell{Temporal \\ Flickering $\uparrow$} & \makecell{Motion \\ Smoothness $\uparrow$} & \makecell{Dynamic \\ Degree $\uparrow$} & \makecell{VBench \\ Mean $\uparrow$} \\
    \midrule
    \textcolor[rgb]{ .502,  .502,  .502}{Ground-Truth} & \textcolor[rgb]{ .502,  .502,  .502}{-} & \textcolor[rgb]{ .502,  .502,  .502}{37} & \textcolor[rgb]{ .502,  .502,  .502}{-} & \textcolor[rgb]{ .502,  .502,  .502}{0.936 } & \textcolor[rgb]{ .502,  .502,  .502}{0.555} & \textcolor[rgb]{ .502,  .502,  .502}{0.712 } & \textcolor[rgb]{ .502,  .502,  .502}{0.980} & \textcolor[rgb]{ .502,  .502,  .502}{0.982} & \textcolor[rgb]{ .502,  .502,  .502}{0.86} & \textcolor[rgb]{ .502,  .502,  .502}{0.84} \\
    GI    & 1.5B  & 25    & 674   & \textbf{0.952} & 0.534  & \textbf{0.699} & 0.959  & 0.985  & 0.43  & 0.76  \\
    ViBidSampler & 1.5B  & 25    & 693   & 0.924  & 0.534  & 0.657  & 0.947  & 0.973  & 0.75  & 0.80  \\
    FCVG  & 1.5B  & 25    & 789   & 0.945  & 0.545  & 0.677  & 0.965  & 0.987  & 0.36  & 0.75  \\
    Framer & 1.5B  & 25    & 702   & 0.933  & 0.487  & 0.640  & 0.969  & 0.986  & 0.47  & 0.75  \\
    Wan2.1-Fun (Wan) & 1.3B  & 37    & 622   & 0.944  & 0.531  & 0.674  & 0.960  & 0.985  & 0.65  & 0.79  \\
    \textbf{Wan+Ours} & 1.3B  & 37    & 561   & 0.937  & 0.529  & 0.680  & 0.968  & \textbf{0.990} & 0.79  & 0.82  \\
    SFI   & 14B   & 33    & 786   & 0.937  & 0.548  & 0.677  & 0.965  & 0.975  & 0.84  & 0.82  \\
    FramePack (FP) & 13B   & 37    & 878   & 0.945  & \textbf{0.573} & 0.683  & 0.965  & 0.983  & 0.64  & 0.80  \\
    FP+finetuned & 13B   & 37    & 617   & 0.939  & 0.565  & 0.654  & 0.972  & 0.989  & 0.79  & 0.82  \\
    \textbf{FP+Ours} & 13B   & 37    & \textbf{477} & 0.949  & 0.549  & 0.652  & \textbf{0.984} & 0.982  & \textbf{0.88} & \textcolor[rgb]{ .753,  0,  0}{\textbf{0.83}} \\
    \midrule
    \textcolor[rgb]{ .502,  .502,  .502}{Ground-Truth} & \textcolor[rgb]{ .502,  .502,  .502}{-} & \textcolor[rgb]{ .502,  .502,  .502}{73} & \textcolor[rgb]{ .502,  .502,  .502}{-} & \textcolor[rgb]{ .502,  .502,  .502}{0.949} & \textcolor[rgb]{ .502,  .502,  .502}{0.554} & \textcolor[rgb]{ .502,  .502,  .502}{0.712} & \textcolor[rgb]{ .502,  .502,  .502}{0.988 } & \textcolor[rgb]{ .502,  .502,  .502}{0.989} & \textcolor[rgb]{ .502,  .502,  .502}{0.73} & \textcolor[rgb]{ .502,  .502,  .502}{0.82} \\
    Wan2.1-Fun (Wan) & 1.3B  & 73    & 412   & 0.943  & 0.528  & 0.679  & 0.974  & 0.987  & 0.51  & 0.77  \\
    \textbf{Wan+Ours} & 1.3B  & 73    & 367   & 0.941  & 0.537  & \textbf{0.687} & 0.983  & 0.988  & 0.56  & 0.78  \\
    SFI   & 14B   & 65    & 499   & 0.933  & 0.542  & 0.669  & 0.973  & 0.988  & 0.75  & 0.81  \\
    FramePack (FP) & 13B   & 73    & 565   & 0.941  & \textbf{0.575} & 0.687  & 0.973  & 0.990  & 0.64  & 0.80  \\
    FP+finetuned & 13B   & 73    & 440   & 0.943  & 0.567  & 0.665  & 0.977  & 0.991  & 0.70  & 0.81  \\
    \textbf{FP+Ours} & 13B   & 73    & \textbf{321} & \textbf{0.949} & 0.557  & 0.680  & \textbf{0.987} & \textbf{0.994} & \textbf{0.76} & \textcolor[rgb]{ .753,  0,  0}{\textbf{0.82}} \\
    \bottomrule
    \end{tabular}%
  \label{tab:base2}%
\end{table*}%

For the 37-frame setting on VidGen-1M, several baselines achieve high scores on isolated metrics but sacrifice overall fidelity. GI leads in subject consistency (0.921) but lags in imaging quality; ViBidSampler attains the lowest FVD (693) yet exhibits poor imaging quality (0.540); SFI achieves the maximum dynamic degree (1.00) but deviates notably from the GT distribution (0.98). In contrast, our \textit{Wan+Ours} and \textit{FP+Ours} both achieve the highest average VBench score (0.81), matching the GT average. Moreover, \textit{FP+Ours} closely aligns with GT on imaging quality (0.602 vs. 0.636), temporal flickering (0.988 vs. 0.949), and motion smoothness (0.992 vs. 0.987), demonstrating more faithful motion dynamics. For the 73-frame setting, \textit{Wan+Ours} achieves the lowest FVD (563), while \textit{FP+Ours} again secures the top average VBench (0.81, matching GT). Notably, FramePack's motion smoothness drops from 0.992 to 0.984 as length increases, whereas \textit{FP+Ours} maintains 0.992, underscoring its robustness against temporal drift.

To further validate generalization, we conduct zero-shot evaluation on UltraVideo. In the 37-frame setting, \textit{FP+Ours} achieves the lowest FVD (477) and highest average VBench (0.83, approaching GT 0.84), with the closest match to GT on dynamic degree (0.88 vs. 0.86) and motion smoothness (0.982 vs. 0.982). GI, despite excelling in imaging quality (0.699), drops significantly in dynamic degree (0.43) and overall average (0.76). For the 73-frame setting, \textit{FP+Ours} again achieves the best FVD (321), highest average VBench (0.82, matching GT 0.82), and superior proximity to GT on dynamic degree (0.76 vs. 0.73) and motion smoothness (0.994 vs. 0.989). These results consistently indicate that our bidirectional training generalizes robustly across diverse video distributions.

\noindent\textbf{Qualitative Comparisons.}
Figure~\ref{fig:baseline} presents the qualitative comparison across different sequence lengths. Among the lightweight baselines, we observe diverse failure patterns. GI and Wan2.1‑Fun generally struggle to preserve fine-grained details from the input frames, often producing blurred or oversmoothed regions. ViBidSampler suffers from severe temporal confusion, manifesting as bidirectional artifacts where motion appears to emanate from both ends simultaneously, breaking the natural temporal order. FCVG generates sequences that lack motion fluidity, exhibiting abrupt jumps and discontinuous transitions between key poses. Framer fails to maintain subject consistency, with the foreground object drifting or morphing across frames. In contrast, our Wan+Ours variant, while not entirely immune to challenges in reconstructing intricate facial features, produces substantially smoother and more coherent motion trajectories than all these lightweight counterparts, demonstrating the effectiveness of our bidirectional regularization.

In the 73-frame setting, FramePack exhibits premature convergence, where the generated video rushes to the end frame without a seamless transition. Similarly, while SFI maintains a high dynamic degree, it often introduces erratic and semantically meaningless motions. In contrast, \textit{FP+Ours} substantially enhances the temporal dynamics of the base FramePack model, successfully generating videos with fluid, coherent transitions from the start frame to the end frame.
The integration of bidirectional training and directional tokens allows our methods to handle complex dynamics more effectively than baselines, resulting in videos that are not only visually high-quality but also temporally stable.

\begin{figure}
 \centering
 \includegraphics[width=0.9\linewidth]{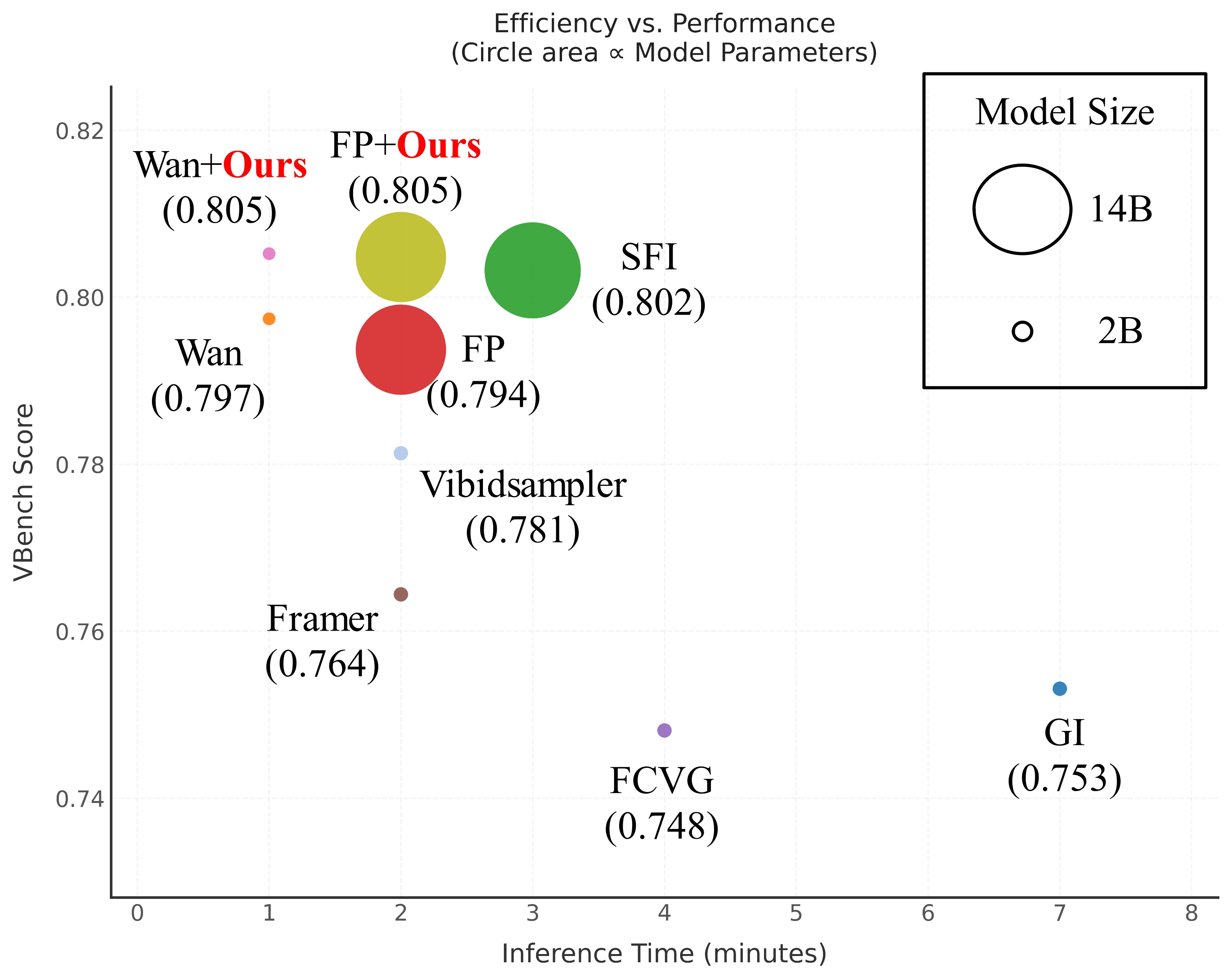}
 \caption{Efficiency vs. Performance. Models closer to the top-left corner exhibit faster inference and higher quality. Circle area indicates model parameter size. Our method outperforms the baselines in VBench score while preserving the same inference efficiency as the original backbones.}
 \label{fig:eff}
  \vspace{-.1in}
\end{figure}

\noindent\textbf{Efficiency.} Figure~\ref{fig:eff} presents a trade-off analysis between inference efficiency and generation quality across baseline methods. All experiments are conducted on an NVIDIA H800 GPU with 80GB of memory to ensure a fair comparison of inference time. \textit{Wan+Ours} and \textit{FP+Ours} preserve the exact inference time of their respective backbones yet deliver superior performance on the quality axis. This confirms that our approach enhances generation capability across all evaluated dimensions without introducing any computational overhead or speed compromise.

\begin{figure}
 \centering
 \includegraphics[width=\linewidth]{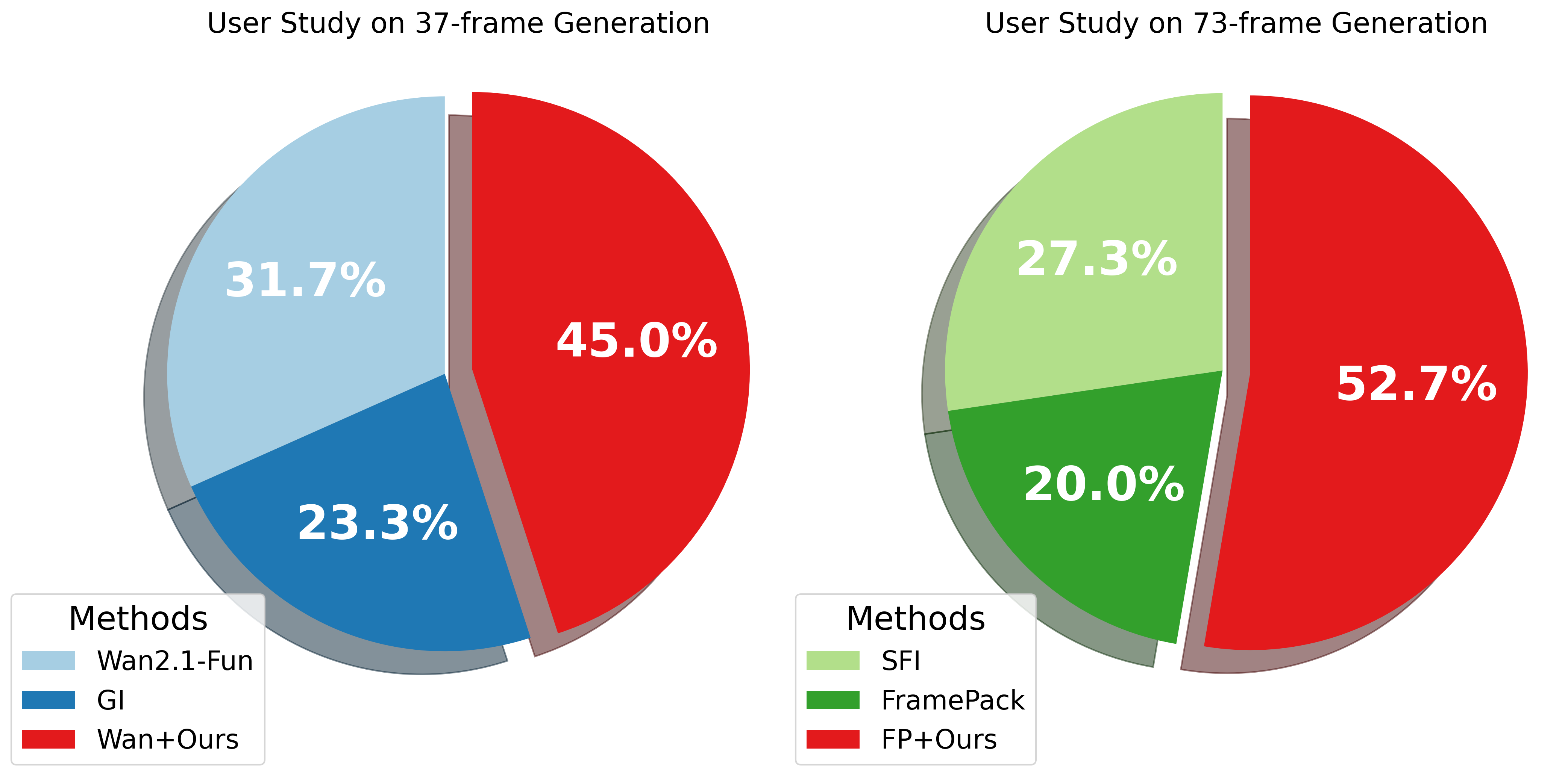}
 \caption{User Study. In both short and long scenarios, our approach is consistently preferred by a large margin, confirming its effectiveness in producing visually appealing and temporally stable videos.}
\label{fig:us}
 \vspace{-.1in}
\end{figure}

\noindent\textbf{User study.} To complement the automatic metrics with human perceptual evaluation, we conduct a user study where 20 participants perform best‑of‑three comparisons. Each trial presents three videos generated by different methods, all sharing identical start/end frames and text prompts. Participants are asked to select the video with the most visually plausible and temporally coherent motion. Each participant completed 15 triplets, choosing one of three options per triplet, resulting in 300 individual votes in total. Results are presented in Figure~\ref{fig:us}. For the 37‑frame setting, we compare GI, Wan2.1‑Fun, and \textit{Wan+Ours}. Our method is selected as the best in 45\% of the trials, substantially outperforming the other two baselines. For the 73‑frame setting, we compare SFI, FramePack, and \textit{FP+Ours}. Here, our method achieves an even higher preference rate of 52.7\%. These results confirm that our bidirectional cycle‑consistent training yields videos that are not only quantitatively competitive but also more appealing to human viewers, especially for long‑range interpolation where temporal coherence is most critical. The strong preference for our method across both short and long sequences underscores its practical advantage in real‑world applications.

\begin{table*}
  \centering
  \caption{Quantitative ablation study of key components in our framework. We evaluate all variants on 37-frame and 73-frame interpolation task using FVD (lower is better) and five VBench metrics (higher is better). Our full model achieves the best overall performance, with consistent improvements in imaging quality and dynamic degree.}
    \begin{tabular}{c|c|c|c|c|c|c|c|c|c}
    \toprule
    Frames & Setting & \multicolumn{1}{c|}{FVD $\downarrow$} &\makecell{Subject \\ Consistency $\uparrow$} & \makecell{Aesthetic \\ Quality $\uparrow$} & \makecell{Imaging \\ Quality $\uparrow$} & \makecell{Temporal \\ Flickering $\uparrow$} & \makecell{Motion \\ Smoothness $\uparrow$} & \makecell{Dynamic \\ Degree $\uparrow$} & \makecell{VBench \\ Mean $\uparrow$} \\
    \midrule
    \multirow{6}[2]{*}{37} & w/o Reverse Training & 937   & 0.876  & 0.466  & 0.588  & 0.987  & \textbf{0.992} & 0.58  & 0.75  \\
          & Forward Data Augmentation & 932   & 0.880  & 0.461  & 0.553  & \textbf{0.988}  & 0.991  & 0.59  & 0.74  \\
          & w/o Directional Tokens & 1018  & \textbf{0.899} & \textbf{0.470} & 0.590  & 0.975  & 0.990  & 0.67  & 0.77  \\
          & w/o Pixel Loss & 897   & 0.866  & 0.462  & 0.548  & 0.964  & 0.982  & 0.89  & 0.79  \\
          & Mixed Training Length & 901   & 0.883  & 0.466  & 0.594  & 0.977  & 0.988  & 0.87  & 0.80  \\
          & Full model & \textbf{885} & 0.885  & 0.465  & \textbf{0.602} & \textbf{0.988} & \textbf{0.992} & \textbf{0.90} & \textcolor[rgb]{ .753,  0,  0}{\textbf{0.81}} \\
    \midrule
    \multirow{6}[2]{*}{73} & w/o Reverse Training & 760   & 0.876  & 0.474  & 0.596  & \textbf{0.993} & \textbf{0.995} & 0.27  & 0.70  \\
          & Forward Data Augmentation & 706   & 0.875  & 0.473  & 0.575  & 0.989  & 0.986  & 0.36  & 0.71  \\
          & w/o Directional Tokens & 712   & 0.878  & 0.473  & 0.598  & 0.977  & 0.991  & 0.69  & 0.77  \\
          & w/o Pixel Loss & 628   & 0.873  & 0.471  & 0.565  & 0.970  & 0.986  & \textbf{0.87} & 0.79  \\
          & Mixed Training Length & 626   & 0.878  & 0.474  & 0.620  & 0.987  & 0.992  & 0.63  & 0.76  \\
          & Full model & \textbf{601} & \textbf{0.882} & \textbf{0.477} & \textbf{0.632} & 0.987  & 0.992  & \textbf{0.87} & \textcolor[rgb]{ .753,  0,  0}{\textbf{0.81}} \\
    \bottomrule
    \end{tabular}%
  \label{tab:ab}%
   \vspace{-.1in}
\end{table*}%

\subsection{Ablation Studies and Further Discussion}
\label{sec:exp_cycle}
To systematically dissect the contribution of each component, we design five ablation variants trained on the same dataset and backbone with identical hyperparameters. (a) \textit{w/o Reverse Training} removes backward generation to isolate the effect of bidirectional cycle-consistent supervision. (b) \textit{Forward Data Augmentation} serves as a control: we augment the backward sequence as an additional forward sample without reversing its direction, distinguishing cycle-consistency from mere data volume increase. (c) \textit{w/o Directional Tokens} disables learnable directional tokens and uses a shared prompt for both directions, testing whether explicit orientation cues are necessary. (d) \textit{w/o Pixel Loss} removes pixel-level reconstruction loss and relies solely on latent-space supervision, evaluating the importance of multi-scale supervision for visual detail preservation. (e) \textit{Mixed Training Length} replaces our two-stage curriculum with joint training on both short and long sequences from the start, assessing the effectiveness of progressive curriculum learning.

As shown in Table~\ref{tab:ab}, quantitative results reveal distinct roles of each component. Variant (a) without reverse training causes dynamic degree to drop from 0.90 to 0.58 (37 frames) and 0.87 to 0.27 (73 frames), confirming that unidirectional training collapses to low-motion solutions. Variant (b) with forward data augmentation yields only marginal FVD improvements (932 vs. 937, 706 vs. 760), proving that explicit reversal symmetry, not data volume, is key. Variant (c) without directional tokens suffers from higher FVD (1018 vs. 885, 712 vs. 601) and worse temporal flickering (0.975 vs. 0.988), indicating that orientation cues are essential for disambiguation. Variant (d) omitting pixel loss achieves competitive dynamic degree (0.89 and 0.87) but degrades imaging quality (0.548 and 0.565) and flickering (0.964 and 0.970), highlighting the need for multi-scale supervision. Variant (e) with mixed training length underperforms the full model with higher FVD (901 vs. 885, 626 vs. 601), validating our progressive curriculum.
Overall, our full model achieves the lowest FVD and the highest average VBench score across both settings, with consistent improvements in imaging quality and dynamic degree, which underscores the effectiveness of our bidirectional cycle‑consistent training paradigm.

\begin{figure*}[t]
 \centering
 \includegraphics[width=0.8\linewidth]{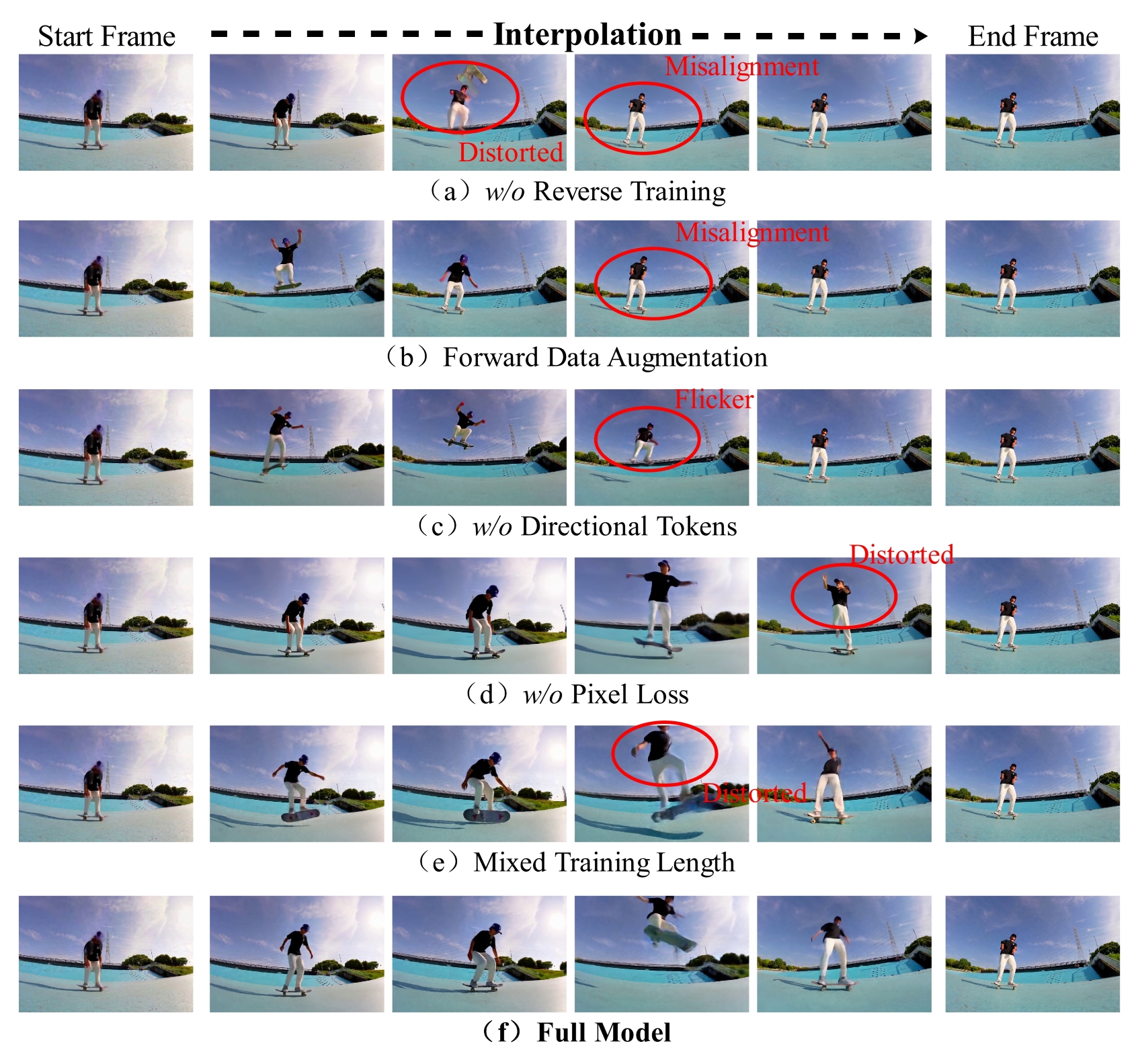}
  \vspace{-.1in}
 \caption{Qualitative ablation study of key components in our framework. Our full model generates a fluid motion sequence that naturally evolves from a ``sliding'' preparation into a full ``jump'', ensuring high temporal coherence and dynamic realism. Videos can be viewed in our project website.}
 \label{fig:ab}
 \vspace{-.1in}
\end{figure*}

Qualitative results are shown in Figure~\ref{fig:ab}.
Variant (a), \textit{w/o Reverse Training}, produces a severely degraded video that collapses prematurely: the subject jumps abruptly to the final pose within the first few frames and remains static for the remainder of the sequence, lacking any meaningful transition or progressive motion. 
Variant (b), Forward Data Augmentation, generates a sequence that captures only a simplistic "jump" motion, omitting the preparatory "sliding" phase. 
Variant (c), w/o Directional Tokens, also produces only a basic jump motion, but with a notable degradation in object persistence. In several intermediate frames, the skateboard disappears entirely and reappears only in the final pose, creating a visually jarring discontinuity.
Variant (d), w/o Pixel Loss, produces a sequence that generally follows the intended motion direction, but the jumping phase itself lacks fluidity and natural dynamics. The subject's transition from the sliding preparation into the jump appears abrupt and mechanically stiff.
Variant (e), Mixed Training Length, yields a disjointed sequence with noticeable motion discontinuities. The action appears fragmented, creating an unnatural and jerky visual experience.
In stark contrast, our full model successfully synthesizes a fluid and semantically meaningful trajectory. The subject seamlessly transitions from the initial "sliding" phase into the subsequent "jump", with each intermediate frame contributing to a smooth and continuous motion path.

\begin{figure}
 \centering
 \includegraphics[width=0.9\linewidth]{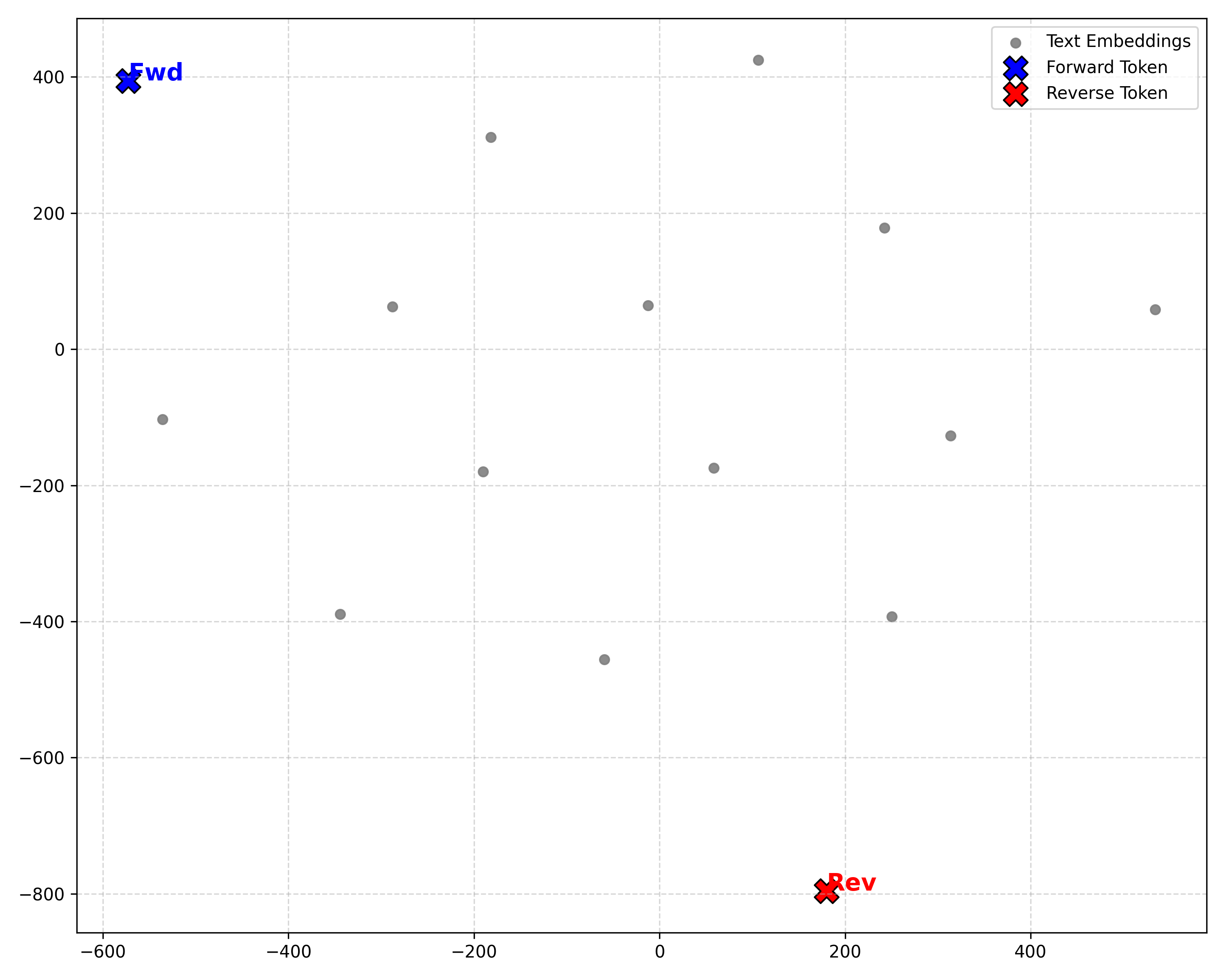}
 \caption{T-SNE visualization of directional tokens. The forward and backward tokens appear as two isolated points, located far from the dense cluster of standard text embeddings. Their near-orthogonal relationship (cosine similarity $= 0.0221$) confirms that our model decouples temporal direction from semantic content.}
\label{fig:tsne}
 \vspace{-.1in}
\end{figure}

\noindent\textbf{Directional token analysis.}
As illustrated in Figure~\ref{fig:tsne}, the learned Forward and Reverse tokens occupy distinct, orthogonal regions in the latent space, far from the cluster of standard text embeddings. Quantitatively, the cosine similarity between these two tokens is merely 0.0221, indicating an angle close to $90^\circ$. This orthogonality suggests that our method successfully decouples temporal directionality from semantic content. Instead of treating ``reverse generation" as a simple negation of the forward process, the model learns it as an independent control dimension, allowing for precise temporal manipulation without compromising semantic consistency. We have included videos in the supplementary material showing results from forward and reverse tokens on paired inputs.

\noindent\textbf{Effectiveness of our bidirectional training.} We evaluate both forward and backward generation directions, with backward videos temporally reversed before evaluation against the forward ground-truth. As shown in Table~\ref{tab:bw}, our forward generation substantially outperforms the untrained baseline across two datasets, confirming the benefits of our training framework. Interestingly, the reversed backward generation also achieves competitive results, indicating that the model genuinely learns meaningful reverse dynamics. However, its performance slightly lags behind the forward direction, which is expected given the inherent difficulty of predicting the past from the future. Most importantly, the primary goal of incorporating backward supervision is not to produce high-quality reverse videos per se, but to regularize the forward generation through cycle-consistency constraints. 


\begin{table}
  \centering
  \caption{Evaluation of forward and backward generation.
  $\hat{\mathcal{V}}_{\mathrm{fwd}}$ and $\hat{\mathcal{V}}_{\mathrm{bwd}}$
  denote the generated forward and backward sequences; backward videos
  are temporally reversed before evaluation against the forward ground
  truth. $\mathcal{C}_{\ell_1}$ is the per-pixel mean absolute error
  between $\hat{\mathcal{V}}_{\mathrm{fwd}}$ and
  $\mathrm{Reverse}(\hat{\mathcal{V}}_{\mathrm{bwd}})$ of the same input
  pair (pixel range $[0,1]$), serving as an empirical measure of the
  cycle discrepancy in Eq.~\ref{eq:cycle}. For the baseline, backward
  generation is obtained by swapping the endpoint conditions.}
  \resizebox{0.99\linewidth}{!}{
    \begin{tabular}{c|c|c|ccc|ccc}
    \toprule
    \multirow{2}{*}{Frames} & \multirow{2}{*}{Methods} & \multirow{2}{*}{Videos}
      & \multicolumn{3}{c|}{VidGen-1M} & \multicolumn{3}{c}{UltraVideo} \\
      & & & FVD $\downarrow$ & VBench $\uparrow$ & $\mathcal{C}_{\ell_1} \downarrow$
          & FVD $\downarrow$ & VBench $\uparrow$ & $\mathcal{C}_{\ell_1} \downarrow$ \\
    \midrule
    \multirow{4}{*}{37}
      & \multirow{2}{*}{Baseline}
        & $\hat{\mathcal{V}}_{\mathrm{fwd}}$                   & 1049 & 0.79 & \multirow{2}{*}{0.141} & 878 & 0.80 & \multirow{2}{*}{0.094} \\
      & & $\mathrm{Reverse}(\hat{\mathcal{V}}_{\mathrm{bwd}})$ & 1075 & 0.76 &  & 910 & 0.79 & \\
    \cmidrule{2-9}
      & \multirow{2}{*}{Ours}
        & $\hat{\mathcal{V}}_{\mathrm{fwd}}$                   & \textbf{885} & \textbf{0.81} & \multirow{2}{*}{\textbf{0.108}} & \textbf{477} & \textbf{0.83} & \multirow{2}{*}{\textbf{0.057}} \\
      & & $\mathrm{Reverse}(\hat{\mathcal{V}}_{\mathrm{bwd}})$ & 987 & 0.77 &  & 625 & \textbf{0.83} & \\
    \midrule
    \multirow{4}{*}{73}
      & \multirow{2}{*}{Baseline}
        & $\hat{\mathcal{V}}_{\mathrm{fwd}}$                   & 686 & 0.79 & \multirow{2}{*}{0.132} & 565 & 0.80 & \multirow{2}{*}{0.086} \\
      & & $\mathrm{Reverse}(\hat{\mathcal{V}}_{\mathrm{bwd}})$ & 700 & 0.77 &  & 677 & 0.76 & \\
    \cmidrule{2-9}
      & \multirow{2}{*}{Ours}
        & $\hat{\mathcal{V}}_{\mathrm{fwd}}$                   & \textbf{601} & \textbf{0.81} & \multirow{2}{*}{\textbf{0.089}} & \textbf{477} & \textbf{0.82} & \multirow{2}{*}{\textbf{0.051}} \\
      & & $\mathrm{Reverse}(\hat{\mathcal{V}}_{\mathrm{bwd}})$ & 645 & 0.80 &  & 625 & 0.81 & \\
    \bottomrule
    \end{tabular}%
  }
  \label{tab:bw}%
\end{table}

\begin{figure}
 \centering
 \includegraphics[width=0.9\linewidth]{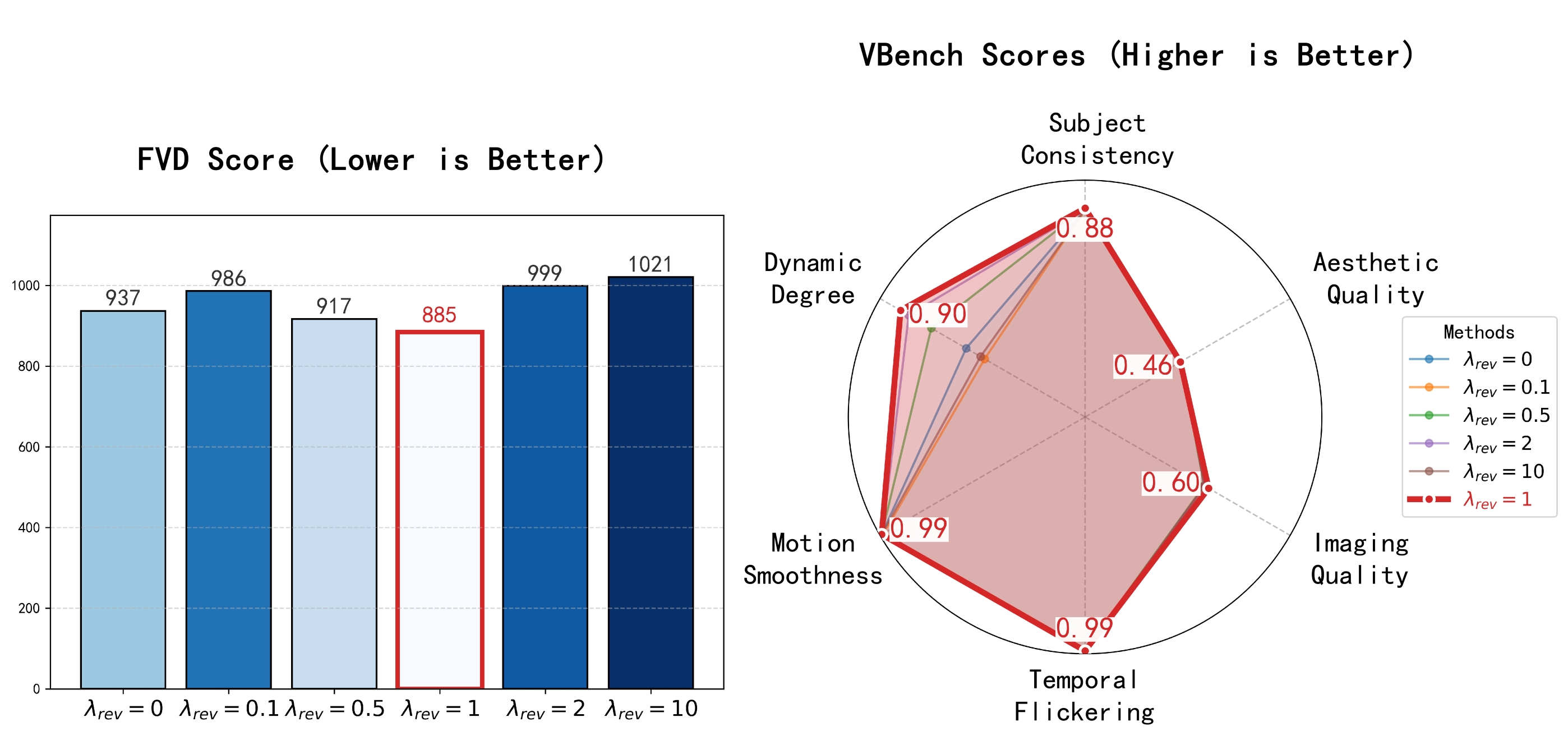}
 \caption{Effect of $\lambda_{\text{rev}}$ on generation quality. $\lambda_{\text{rev}}=0$ yields static videos, $\lambda_{\text{rev}}=1$ achieves the best dynamic degree (0.90), while $\lambda_{\text{rev}}=10$ over-constrains and degrades performance.}
 \label{fig:rw}
  \vspace{-.1in}
\end{figure}

\noindent\textbf{Ablation Study on Reverse Loss Weight.}
\noindent To understand the impact of the reverse consistency loss, we ablate the weighting coefficient $\lambda_{\text{rev}}$ across $\{0, 0.1, 0.5, 1, 2, 10\}$, with results shown in Figure~\ref{fig:rw}. The \textit{Dynamic Degree} exhibits a clear trend: at $\lambda_{\text{rev}}=0$, the model produces static videos with a score of 0.58; as $\lambda_{\text{rev}}$ increases, the dynamic degree improves, peaking at 0.90 with $\lambda_{\text{rev}}=1$. However, over-constraining with $\lambda_{\text{rev}}=10$ collapses the score to 0.51, indicating that excessive backward supervision restricts generative diversity. We therefore adopt $\lambda_{\text{rev}}=1$ as the default setting for all experiments.

\noindent\textbf{Limitation.} Our bidirectional training framework implicitly assumes approximate reversibility in the underlying dynamics, which may not hold for all real-world processes, particularly those involving stochastic transitions, information dissipation, or irreversible physical changes. In such cases, enforcing strict cycle consistency could potentially bias the model toward overly smoothed or physically inaccurate trajectories (\eg, systems akin to ``Schrödinger's cat'' where the act of observation collapses the state).

\section{Conclusion}
This paper is motivated by a simple yet critical observation: existing video diffusion models struggle with temporal retrospection, such as predicting frames in reverse. Leveraging this insight, we present a bidirectional framework for text-guided video frame interpolation that leverages time-reversal symmetry to enhance long-range temporal coherence. By introducing learnable directional tokens and a symmetric training objective within a shared backbone, our method effectively resolves motion ambiguity without architectural modifications. Extensive experiments confirm that our approach achieves state-of-the-art visual fidelity and motion smoothness while maintaining the high inference efficiency of the base model. These results highlight the potential of cycle-consistent supervision in generative video modeling. Looking forward, we aim to incorporate explicit geometric constraints to further improve the physical realism of our method in highly dynamic environments.



 

\bibliographystyle{IEEEtran}
\bibliography{main}

\vfill

\end{document}